\PassOptionsToPackage{table}{xcolor}
\documentclass{article} 
\usepackage{nips15submit_e,times}
\usepackage{hyperref}
\usepackage{url}
\usepackage{multicol}
\usepackage{algorithm}
\usepackage{algpseudocode}
\usepackage{xcolor}
\usepackage{lipsum}
\usepackage{multicol}
\usepackage[round]{natbib}
\usepackage{enumitem}
\usepackage{graphicx,changepage,amsmath,subcaption}

\newcommand{\sysname}[1]{\textsc{#1}}

\newcommand{\uc}[1]{\protect\uppercase{#1}}
\newtheorem{theorem}{Theorem}

\usepackage{mathtools}
\DeclareMathOperator*{\argmax}{arg\,max}
\DeclareMathOperator*{\diag}{diag}

\title{Progressive EM for Latent Tree Models and Hierarchical Topic Detection}

\author{
Peixian Chen\thanks{http://peixianc.me} \\
Department of Computer Science\\
The Hong Kong University of Science and Technology\\
\texttt{pchenac@cse.ust.hk} \\
\And
Nevin L. Zhang\\
Department of Computer Science\\
The Hong Kong University of Science and Technology\\
\texttt{lzhang@cse.ust.hk} \\
\AND
Leonard K.M. Poon \\
Department of Mathematics and Information Technology\\
The Hong Kong Institute of Education \\
\texttt{kmpoon@ied.edu.hk} \\
\And
Zhourong Chen \\
Department of Computer Science\\
The Hong Kong University of Science and Technology\\
\texttt{zchenbb@cse.ust.hk} \\
}

\nipsfinalcopy 

\begin{document}

\maketitle

\begin{abstract}
\emph{Hierarchical latent tree analysis (HLTA)} is recently proposed as a new method for topic detection. It differs fundamentally from the LDA-based methods in terms of topic definition, topic-document relationship, and learning method. It has been shown to discover significantly more coherent topics and better topic hierarchies.  However, HLTA relies on the \emph{Expectation-Maximization (EM)} algorithm for parameter estimation and hence is not efficient enough to deal with large datasets. In this paper, we propose a method to drastically speed up HLTA using a technique inspired by recent advances in the moments method.  Empirical experiments show that our method  greatly improves the efficiency of HLTA.  It is as efficient as the state-of-the-art LDA-based method for hierarchical topic detection and finds substantially better topics and topic hierarchies.
\end{abstract}

\section{\uc{Introduction}}
\vspace{-2mm}
Detecting topics and topic hierarchies from document collections, along with its many potential applications, is a major research area in Machine Learning. Currently the predominant approach to topic detection is \emph{latent Dirichlet allocation} (LDA)~\citep{blei03latent}. LDA has been developed  to detect topics and to model relationships among them, including topic correlations~\citep{blei2007correlated}, topic hierarchies~\citep{blei10nested,paisley2012nested}, and topic evolution~\citep{blei2006dynamic}. We collectively name these methods {\em LDA-based methods}.  In those methods,  a topic is a probability distribution over a vocabulary and a document is a mixture of topics. Therefore LDA is a type of \emph{mixture membership model}.

A totally different approach to hierarchical topic detection is recently proposed by \cite{liu14hierarchical}. It is called \emph{hierarchical latent tree analysis} (HLTA), where topics are organized hierarchically as a \emph{latent tree model} (LTM)~\citep{zhang04hierarchical,zhang08discovery} such as the one in Fig~\ref{fig.toy-hltm}.  In HLTA, a topic is a state of a latent variable and it corresponds to a collection of documents, and a document can belong to multiple topics. HLTA is therefore a type of \emph{multiple membership model}.

Empirical results from \cite{liu14hierarchical} indicate that HLTA finds significantly better topics and topic hierarchies than \emph{hierarchical latent Dirichlet allocation} (hLDA), the first LDA-based method for hierarchical topic detection. However, HLTA does not scale up well. It took, for instance, 17 hours to process a NIPS dataset that consists of fewer than 2,000 documents over 1,000 distinct words~\citep{liu14hierarchical}.(Note that hLDA took even longer time.) 

The computational bottleneck of HLTA lies in the use of  the EM algorithm~\citep{dempster77maximum} for parameter estimation. In this paper, we propose \emph{progressive EM} (PEM) as a replacement of EM so as to scale up HLTA. PEM is motivated by the moments method, where parameters are determined by solving equations, each of which involves a small number of model parameters related to two or three observed variables~\citep{chang96full,anandkumar12spectral}. Similarly, PEM works in steps and, at each step, it focuses on a small part of the model parameters and involves only three or four observed variables.

Our new algorithm is hence named PEM-HLTA. It is drastically faster than HLTA. PEM-HLTA finished  processing the aforementioned NIPS dataset within 4 minutes. It only took around 11 hours, on a single desktop computer,  to analyze a version of New York Times dataset that consists of 300,000 articles with 10,000 distinct words. PEM-HLTA is also as efficient as nHDP~\citep{paisley2012nested}, a state-of-the-art LDA-based method for hierarchical topic detection, and it significantly outperforms nHDP, as well as hLDA, in terms of the quality of topics and topic hierarchies.

\vspace{-3mm}

\section{\uc{Preliminaries}\label{sec:pre}}
\vspace{-2mm}
A \emph{latent tree model} (LTM) is a Markov random field over an undirected tree, where the leaf nodes represent observed variables and the internal nodes represent latent variables~\citep{zhang08blatent}. In this paper we assume all variables have finite \emph{cardinality}, i.e., finite number of possible states.

Parameters of an LTM consist of potentials associated with edges and nodes such that the product of all potentials is a joint distribution over all variables. We pick the potentials as follows:  Root the model at an arbitrary latent node, direct the edges away from the root, and specify a marginal distribution for the root and a conditional distribution for each of the other nodes given its parent. Then in Fig~\ref{fig.ud-test}(b), if $Y$ is the root,  the parameters are the distributions $P(Y)$, $P(A\mid Y), P(Z\mid Y), P(C\mid Z)$ and so forth. Because of the way the potentials are picked, LTMs are technically tree-structured Bayesian networks~\citep{pearl88probabilistic}.

LTMs with a single latent variables are known as {\em latent class models (LCMs)}\citep{bartholomew99latent}. They are a type of finite mixture models for discrete data. For example, the model $m_1$ in Fig~\ref{fig.ud-test}(a) defines the following mixture distribution over the observed variables:
\begin{equation}
\begin{small}
\begin{aligned}
P(A,\cdots\!,E) \!=\!\sum\nolimits_{i=1}^{|Y|}P(Y\!=\!i)P(A,\cdots\!,E\mid \!Y\!=\!i) 
\end{aligned}
\end{small}
\end{equation}
where $|Y|$ is the cardinality of $Y$. If the model is learned from a dataset, then the data are partitioned into $|Y|$ soft clusters, each represented by a state of $Y$.
The model $m_2$ in Fig~\ref{fig.ud-test}(b) has two latent variables. Its joint distribution can be reduced to two different but related mixture distributions:
\begin{equation}
\begin{small}
\begin{aligned}
P(A,\cdots\!,E) \!=\!\sum\nolimits_{i=1}^{|Y|}P(Y\!=\!i)P(A,\cdots\!,E\mid \!Y\!=\!i), \nonumber \\
P(A,\cdots\!,E) \!=\! \sum\nolimits_{i=1}^{|Z|}P(Z\!=\!i)P(A,\cdots\!,E\mid \!Z\!=\!i).
\end{aligned}
\end{small}
\end{equation}
The model gives two different ways of partitioning  the data, one represented by $Y$ and the other by $Z$. Hence LTMs are a tool for \emph{multidimensional clustering}~\citep{chen12model}.
\begin{minipage}[t]{\textwidth}
\begin{minipage}{0.6\textwidth}
{
\begin{figure}[H]
\begin{center}
\includegraphics[width=9.2cm]{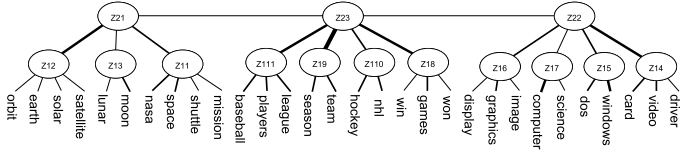}

\caption{\small Latent tree model obtained from a toy text dataset.}

\label{fig.toy-hltm}
\end{center}
\end{figure}
}
\end{minipage}
\qquad
\begin{minipage}{.4\textwidth}
{
\begin{figure}[H]
\begin{center}
	\begin{subfigure}[b]{.55\linewidth}
    \centering
	\includegraphics[scale=.16]{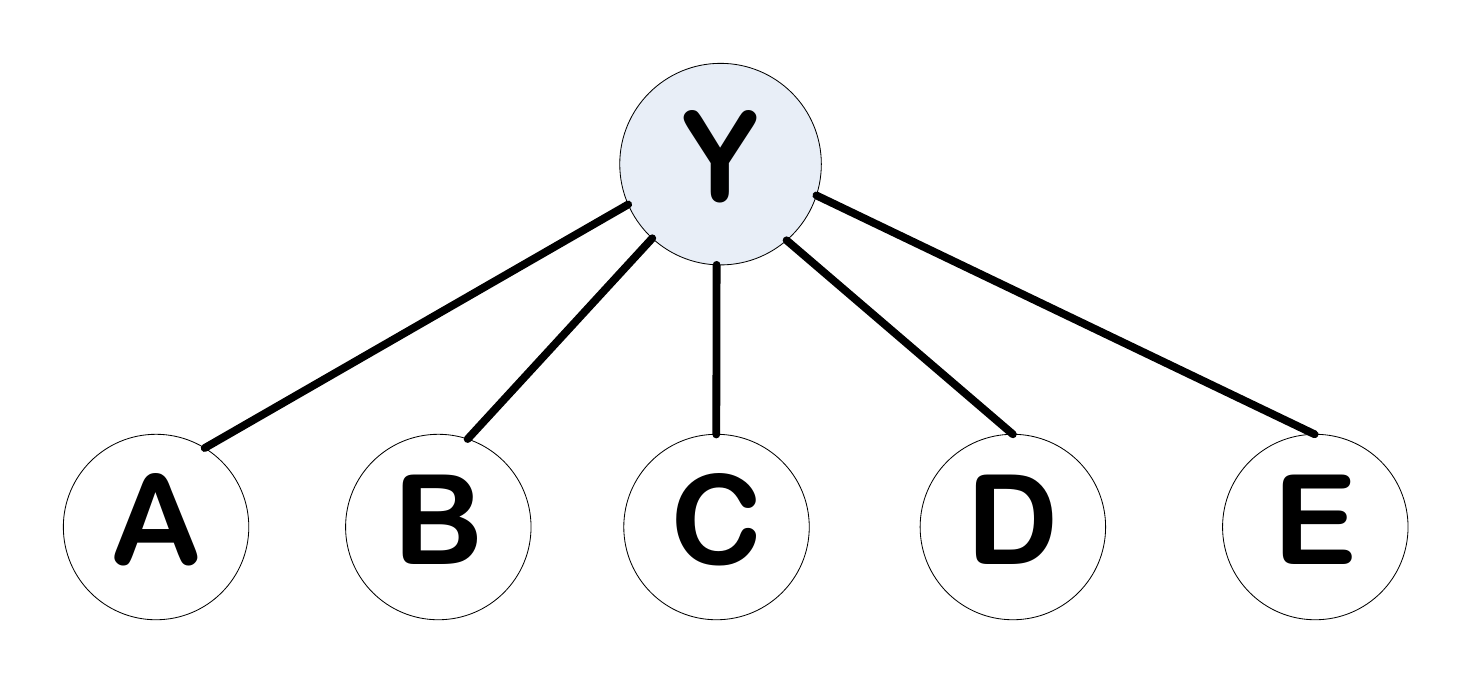}
    \caption{$m_1$}
    \end{subfigure}%
    \begin{subfigure}[b]{.5\linewidth}
    \centering
	\includegraphics[scale=.16]{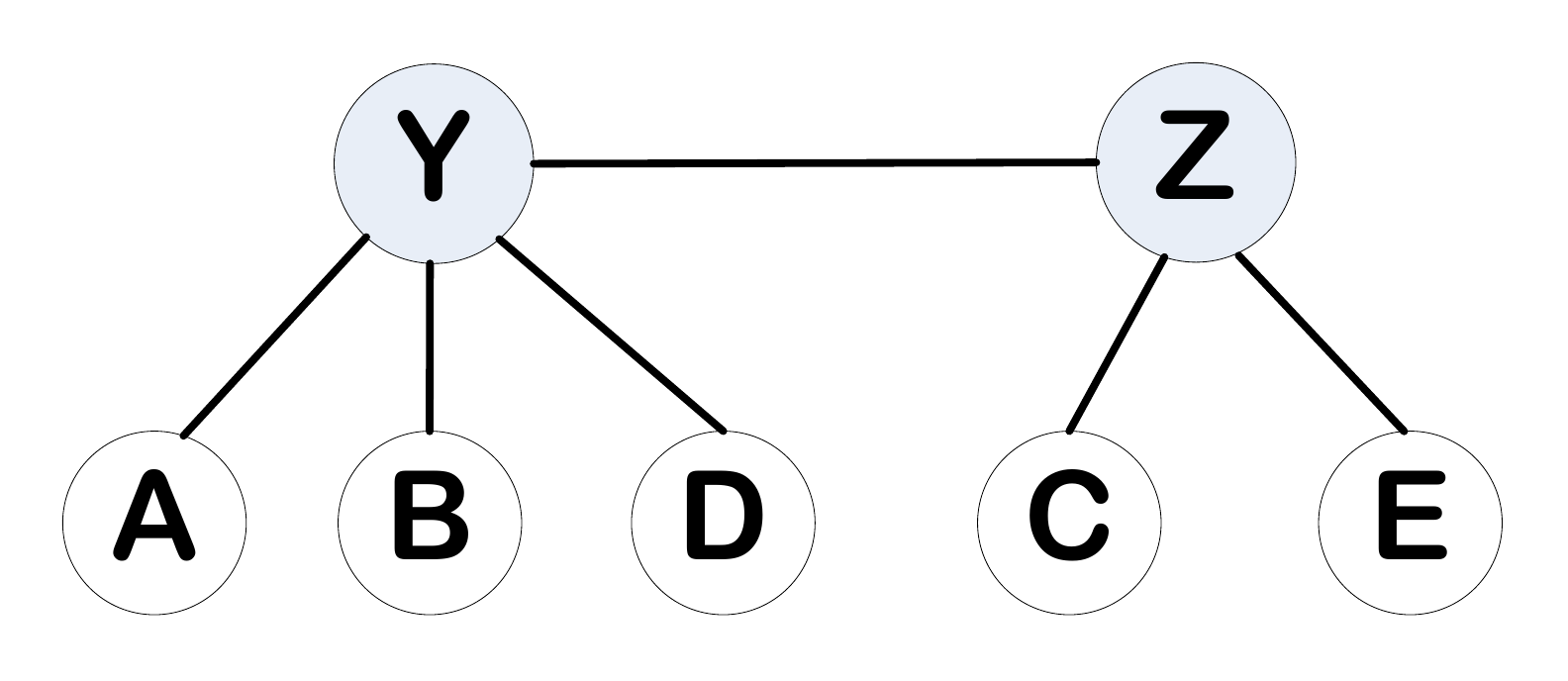}	
    \caption{$m_2$}
    \end{subfigure}

\caption{\small Leaf nodes are observed while others are latent.}\label{fig.ud-test}
\vspace{-2mm}
\end{center}
\end{figure}
}
\end{minipage}
\end{minipage}

\section{\uc{The Algorithm}} \label{sec.algorithm}
\vspace{-2mm}
The input to our PEM-HLTA algorithm is a collection $\mathcal{D}$ of documents, each represented as a binary vector over a vocabulary $\mathcal{V}$. The values in the vector indicate the presence or absence of words in the document.  The output is an LTM, such as the one shown in Fig~\ref{fig.toy-hltm}, where the word variables are at the bottom and the latent variables, all assumed binary, form several levels of hierarchy on top. Each state of a latent variable corresponds to a cluster of documents and is interpreted as a topic. The top level control of PEM-HLTA is given in Algorithm~\ref{alg.pem-hltm}, and subroutines in Algorithms~\ref{alg.buildislands}--\ref{alg.firstisland}.

\subsection{\uc{Top Level Control}}
\vspace{-2mm}

We illustrate the top level control using the example model in Fig~\ref{fig.toy-hltm}, which is learned from a dataset with 30 word variables. In the first pass through the  loop,  the subroutine \sysname{BuildIslands} is called (line~\ref{alg1.buildisland}). It partitions all variables into 11 clusters (Fig~\ref{fig.toy-islands} bottom), which are {\em uni-dimensional} in the sense that the co-occurrences of words in each cluster can be properly modeled using a single latent variable.
A latent variable is introduced for each cluster to form an LCM. We  metaphorically refer to the LCMs as islands and the latent variables in them as level-1 latent variables.

The next step is to link up the 11 islands (line~\ref{alg1:bridge}).
This is done by estimating the {\emph{mutual information} (MI)} \citep{cover2012elements} between every pair of latent variables and building a Chow-Liu tree~\citep{chow68approximating} over them, so as to form an overall model~\citep{liu13greedy}. The result is the model  at the middle of Fig~\ref{fig.toy-islands}.

{
\begin{figure*}[t]
\vspace{-2mm}
\begin{center}
\includegraphics[width=9.5cm]{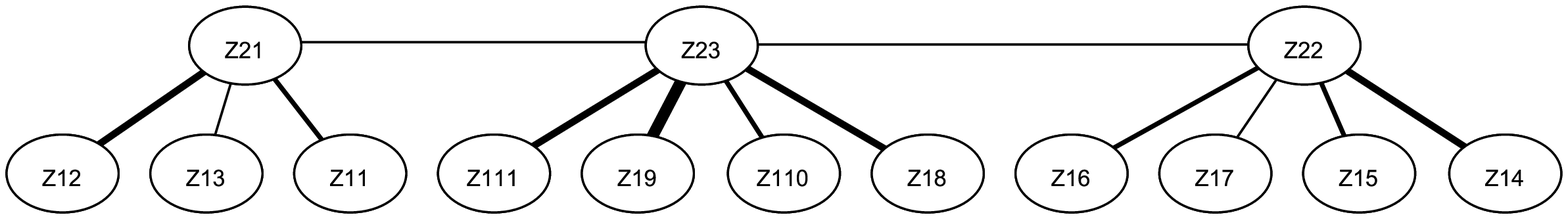} \\ \ \\
\vspace{-2mm}
\includegraphics[width=10cm]{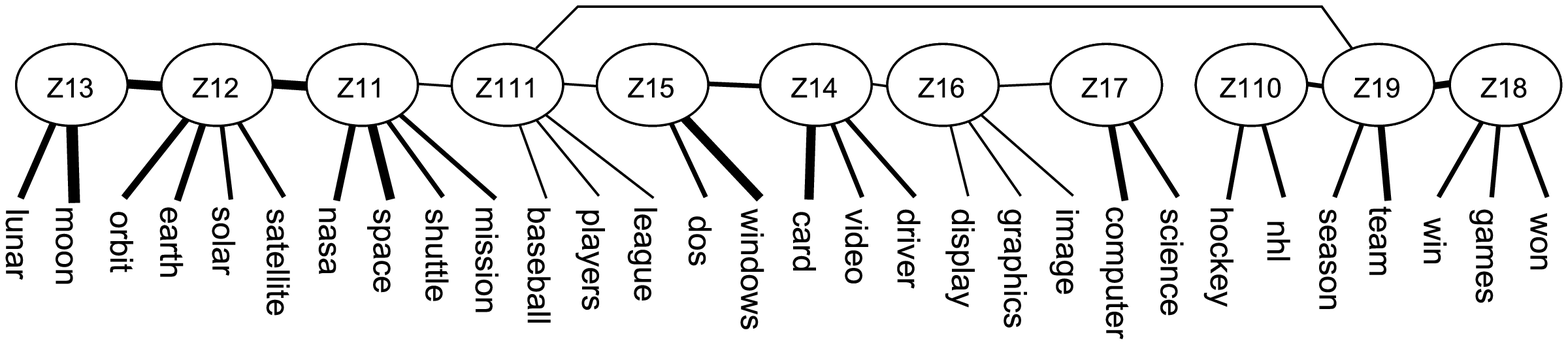} \\  \ \\
\vspace{-3mm}
\includegraphics[width=10cm]{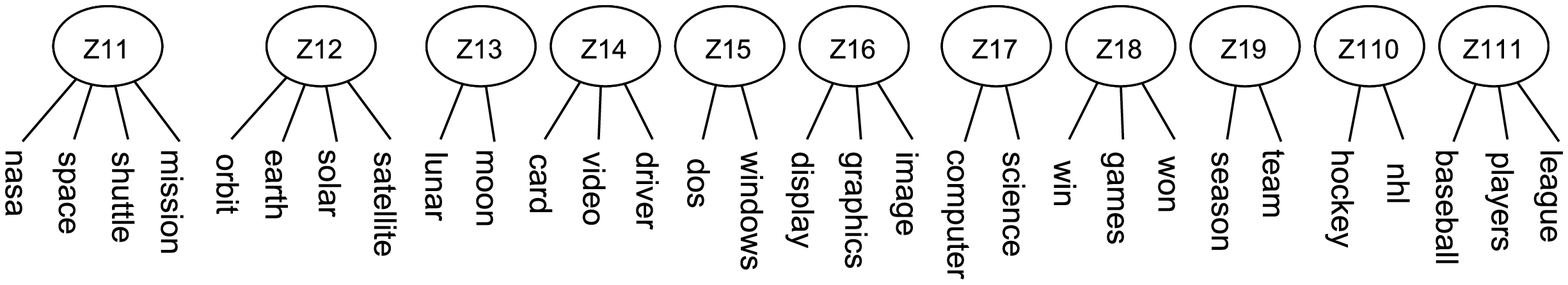}
\end{center}
\vspace{-5mm}
\caption{Intermediate models created by \sysname{PEM-HLTA} on a toy data set.}
\label{fig.toy-islands}
\vspace{-2mm}
\end{figure*}
}
In the subroutine \sysname{HardAssignment}, inference is carried out to compute the posterior distribution of each latent variable for each document.  The document is assigned to the state with the maximum posterior probability.  This results in a dataset over the level-1 latent variables (line~\ref{alg1.hardassign}).  In the second pass through the loop, the level-1 latent variables are partitioned into 3 groups and 3 islands are created. The islands are linked up to form the model shown at the top of Fig~\ref{fig.toy-islands}.
At line~\ref{alg1.stackmodel}, the model at the top of Fig~\ref{fig.toy-islands}  ($m_1$) is stacked on the model in the middle  ($m$)
to give rise to the final model in Fig~\ref{fig.toy-hltm}. While doing so, the subroutine \sysname{StackModels} cuts off the links among the level-1 latent variables. The number of nodes at the top level is below the threshold $\tau$, if we set $\tau=5$, and hence the loop is exited. EM is run on the final model for $\kappa$ steps to improve its parameters (line~\ref{alg1.em}). In our experiments, we set $\kappa=50$. In Section~\ref{sec.TopicExtract} we will discuss how to extract a topic hierarchy from the final model.

\begin{algorithm}[t]
\begin{description}
\small
\item[Inputs:] $\mathcal{D}$---Collection of documents, $\tau$---Upper bound on the number of top-level topics, $\delta$---Threshold used in UD-test, $\kappa$---Number of EM steps on final model.
\item[Outputs:] An HLTM and a topic hierarchy.
\vspace{-2mm}
\end{description}
\small
\begin{multicols}{2}
\begin{algorithmic}[1]
\State $\mathcal{D}_1 \gets \mathcal{D}$, $\mathcal{L} \gets \emptyset$, $m \gets null$.
\Repeat \label{alg1.start}
    \State $\mathcal{L} \gets$ \sysname{BuildIslands}($\mathcal{D}_1$, $\delta$);\label{alg1.buildisland}
    \State $m_1 \gets$  \sysname{BridgeIslands}($\mathcal{L}$, $D_1$);\label{alg1:bridge}

    \If{${m}= null$}
        \State ${m} \gets {m}_1$;
    \Else
        \State $m\gets$ \sysname{StackModels}(${m}_1$, ${m}$); \label{alg1.stackmodel}
    \EndIf
    \State $\mathcal{D}_1 \gets$ \sysname{HardAssignment}($m$, $\mathcal{D}$); \label{alg1.hardassign}

\Until $|\mathcal{L}| < \tau$.  \label{alg1.end}
\State Run EM on $m$ for $\kappa$ steps. \label{alg1.em}
\State \Return $m$ and topic hierarchy extracted from $m$. \label{alg1.return}
\end{algorithmic}
\end{multicols}
\caption{\sysname{PEM-HLTA}($\mathcal{D}$, $\tau$, $\delta$, $\kappa$)}\label{alg.pem-hltm}
\vspace{-4mm}
\end{algorithm}

\subsection{\uc{Building Islands}}
\vspace{-2mm}
The pseudo code for the subroutine \sysname{BuildIslands} is given in
Algorithm~\ref{alg.buildislands}.  It calls another subroutine \sysname{OneIsland}
to identify a uni-dimensional subset of observed variables and builds an LCM with them.
 Then it repeats the process on those observed variables left to create more islands,
 until all variables are included in these islands. Finally, it returns the set of all the islands.

\subsubsection{\uc{Uni-Dimensionality Test}}
\vspace{-2mm}
We rely on the {\em uni-dimensionality test (UD-test)}~\citep{liu13greedy} to determine whether a set $\mathcal{S}$ of variables is uni-dimensional. The idea is to compare two LTMs $m_1$ and $m_2$, where $m_1$  is the best model among all LCMs for $\mathcal{S}$ while $m_2$ is the best model among all LTMs that contain two latent variables. The model selection criterion used is the BIC score ~\citep{schwarz78estimating}.
The set $\mathcal{S}$ is uni-dimensional if the following inequality holds:
\begin{eqnarray}
\mathcal{BIC}(m_2\mid \mathcal{D}) - \mathcal{BIC}(m_1\mid\mathcal{D}) < \delta,
\label{eq.ud-test}
\end{eqnarray}
where $\delta$ is a threshold.  In other words, $\mathcal{S}$ is considered uni-dimensional if the best two-latent variable model is not significantly better than the best one-latent variable model. The quantity on the left hand side of Equation~(\ref{eq.ud-test}) is a large sample approximation of the natural logarithm of Bayes factor~\citep{raftery95bayesian} for comparing $m_1$ and $m_2$.  According to the cut-off values for the Bayes factor, we set $\delta=3$ in our experiments.

\begin{algorithm}[t]
\small
\begin{multicols}{2}
\begin{algorithmic}[1]
\State $\mathcal{V} \gets$ variables in $\mathcal{D}$, $\mathcal{M} \gets \emptyset$.
\While {$|\mathcal{V}| > 0$}
    \State $m \gets$ \sysname{OneIsland}($\mathcal{D}$, $\mathcal{V}$, $\delta$);
    \State $\mathcal{M} \gets \mathcal{M} \cup \{m\}$;
    \State $\mathcal{V} \gets $ variables in $\mathcal{D}$ but not in any $m \in \mathcal{M}$;
\EndWhile
\State \Return $\mathcal{M}$.
\end{algorithmic}
\end{multicols}
\caption{\sysname{BuildIslands}($\mathcal{D}$, $\delta$)}
\label{alg.buildislands}
\vspace{-3mm}
\end{algorithm}

\subsubsection{\uc{Building An Island}}
\vspace{-2mm}
Given dataset $\mathcal{D}$ with variables $\mathcal{V}$, the subroutine \sysname{OneIsland}  identifies a uni-dimensional subset of variables and builds an LCM for them. Define the mutual information between a variable $Z$ and a set  $\mathcal{S}$ as ${\text{MI}}(Z, \mathcal{S})=\max_{A \in \mathcal{S}} {\text{MI}}(Z, A)$.   \sysname{OneIsland} maintains a working set $\mathcal{S}$ of observed variables. Initially, $\mathcal{S}$ contains the pair of variables with the highest MI among all pairs, and a third variable that has the highest MI with the pair (line~\ref{alg3.s}).  At line~\ref{alg3.firstlcm}, an LCM is learned for those three variables using the subroutine \sysname{LearnLCM}, which is given in the Appendix along with some other subroutines. Then other variables are added to $\mathcal{S}$ one by one until the UD-test fails. 
\begin{algorithm}[t]
\small
\begin{multicols}{2}
\begin{algorithmic}[1]
\State \textbf{if} $|\mathcal{V}| \leq 3$, $m \gets$ \sysname{LearnLCM}($\mathcal{D}$, $\mathcal{V}$), \textbf{return} $m$.

\State $\mathcal{S} \gets$ three variables in $\mathcal{V}$ with highest MI,  \label{alg3.s}\\
$\mathcal{V}_1 \gets  \mathcal{V} \setminus \mathcal{S}$;
\State $\mathcal{D}_1 \gets$ \sysname{ProjectData}($\mathcal{D}$, $\mathcal{S}$), \\
$m \gets$ \sysname{LearnLCM}($\mathcal{D}_1$, $\mathcal{S}$).\label{alg3.firstlcm}
\Loop
    \State {\scriptsize $X \gets \argmax_{A \in \mathcal{V}_1} MI(A, \mathcal{S})$, \label{alg3.pickout}
    \State $W \gets \argmax_{A \in S} MI(A, X)$}\label{alg3.pick-x},
    \State $\mathcal{D}_1 \gets$ \sysname{ProjectData}$(\mathcal{D}, \mathcal{S} \cup \{X\})$, $\mathcal{V}_1 \gets \mathcal{V}_1 \setminus \{X\}$.
    \State $m_1 \gets$ \sysname{PEM-LCM}$(m, \mathcal{S}, X, \mathcal{D}_1)$.\label{alg3.lcm}
    \State \textbf{if} $|\mathcal{V}_1| = 0$, \Return $m_1$.\label{alg3.return-lcm}
    \State $m_2 \gets$ \sysname{PEM-LTM-2L}($m$, $\mathcal{S} \setminus \{W\}$, $\{W, X\}$, $\mathcal{D}_1$)\label{alg3.ltm}
    \If{$BIC(m_2|\mathcal{D}_1) - BIC(m_1|\mathcal{D}_1) > \delta$}.\label{alg3.compare}
        \State \Return $m_2$ with $W$, $X$ and their parent removed.\label{alg3.return-island}
    \EndIf
    \State $m \gets m_1$, $\mathcal{S} \gets \mathcal{S} \cup \{X\}$.\label{alg3.add-x}
\EndLoop
\end{algorithmic}
\end{multicols}
\caption{\sysname{OneIsland}($\mathcal{D}$, $\mathcal{V}$, $\delta$)}\label{alg.firstisland}
\end{algorithm}
We illustrate this process using Fig~\ref{fig.ud-test}. Suppose $\mathcal{S}$ initially consists of three variables $A$, $B$, $C$.  Let $D$ be the variable that has the maximum MI with $\mathcal{S}$ among all other variables. Suppose the UD-test passes on $\mathcal{S} \cup \{D\}$, then $D$ is added to $\mathcal{S}$.  Next let $E$ be the variable with the maximum MI with S (line~\ref{alg3.pickout}) and the UD-test is performed on $S\cup{E}= \{A, B, C, D, E\}$ (lines~\ref{alg3.pick-x}-\ref{alg3.return-island}). The two models $m_1$ and $m_2$ used in the test is shown in Fig~\ref{fig.ud-test}. For computational efficiency, we do not search for the best structure for $m_2$. Instead, the structure is determined as follows: Pick the variable in $\mathcal{S}$ that has the maximum MI with $E$ (line~\ref{alg3.pick-x}) (let it be $C$), and group it with $E$ in the model (line~\ref{alg3.ltm}).  The model parameters are estimated using the subroutines \sysname{PEM-LCM} and \sysname{PEM-LTM-2L}, which will be explained in the next section. If the test fails, then $C$, $E$ and $Z$ are removed from $m_2$, and what remains in the model, an LCM, is returned. If the test passes, $E$ is added to $\mathcal{S}$ (line~\ref{alg3.add-x}) and the process continues.

\vspace{-2mm}

\section{\uc{Progressive EM for Model Construction}} \label{sec.pem}
\vspace{-2mm}
PEM-HLTA conceptually consists of a model construction phase (lines~\ref{alg1.start}-\ref{alg1.end}) and a parameter estimation phase (line~\ref{alg1.em}). During the first phase, many intermediate models are constructed. In this section, we present a fast method for estimating the parameters of those intermediate models.

\subsection{\uc{Moments Method for Parameter Estimation}}
\vspace{-2mm}
We begin by presenting a property of LTMs that motivates our new method.  A similar property of HMMs was first discovered by ~\cite{chang96full}. We  introduce some notations using $m_1$ of Fig~\ref{fig.ud-test}. Since all variables have the same cardinality, the conditional distribution $P(A|Y)$ can be regarded as a square matrix, which we denote as $P_{A|Y}$.  Similarly,  $P_{AC}$ is the matrix representation of the joint distribution $P(A, C)$. For a value $b$ of $B$, $P_{b|Y}$ is the vector presentation of $P(B{=}b|Y)$ and $P_{AbC}$ the matrix representation of $P(A, B{=}b, C)$.
\begin{theorem}\label{theorem.1}
\textbf{[~\cite{zhang14study}]} Let $Y$ be the latent variable in an LCM and $A, B, C$  be three of the observed variables. Assume all variables have the same cardinality and the matrices $P_{A|Y}$ and  $P_{AC}$  are invertible. Then we have
\begin{eqnarray}
P_{A|Y} \diag(P_{b|Y})P_{A|Y}^{-1} = P_{AbC}P_{AC}^{-1},
\label{eq.moments}
\end{eqnarray}
where $\diag(P_{b|Y})$ is a diagonal matrix with components of $P_{b|Y}$ as the diagonal elements.
\end{theorem}

The equation implies that the model parameters  $P(B{=}b|Y{=}0),\cdots, P(B{=}b|Y{=}|Y|)$ are the eigenvalues of the matrix on the right, and hence can be obtained from the marginal distributions $P_{AbC}P_{AC}$.

Theorem~\ref{theorem.1} can be used to estimate $P(B|Y)$ under two conditions: (1) There is a good fit between the data and model as if the data were generated from the model, and (2) the sample size is sufficiently large.  In this case, the empirical marginal distributions $\hat{P}(A, B, C)$ and $\hat{P}(A, C)$ computed from data are accurate estimates of the distributions $P(A, B, C)$ and $P(A, C)$ of the model. We can use them to form the matrix $P_{AbC}P_{AC}^{-1}$, and determine $P_{B|Y}$ as the eigenvalues of the matrix. This is called the {\em moments method.}  Note that Theorem~\ref{theorem.1} still applies when replacing edges like $(Y,A)$ with paths. For example in Fig~\ref{fig.ud-test}(b), if $P(C|Z)$ and $P(E|Z)$ are to be estimated, a third observed variable can be chosen from $(A,B,D)$ as long as there is path from $Z$ to this observed variable.

Theorem~\ref{theorem.1} can be also used to estimate all the parameters of the model $m_1$ in Fig~\ref{fig.ud-test}. First, we can estimate $P(B|Y)$ using Equation~\ref{eq.moments} in the sub-model $Y$-$\{A, B, C\}$. By swapping the roles of variables, we can also estimate $P(A|Y)$ and $P(C|Y)$ in the sub-model. Next we can consider the sub-model $Y$-$\{B, C, D\}$ and estimate $P(D|Y)$ with $P(B|Y)$ and $P(C|Y)$ fixed. Finally, we can consider the sub-model $Y$-$\{C, D, E\}$ and estimate $P(E|Y)$ there with $P(C|Y)$ and $P(D|Y)$ fixed. Note that the parameters are estimated in steps instead of all at once. Hence we call this scheme {\em progressive parameter estimation}.

\subsection{\uc{Progressive EM}}
\vspace{-2mm}

The moments method is not iterative and hence can be drastically faster than EM. Unfortunately, it does not produce high quality estimates when the model does not fit  data well and/or the sample size is not sufficiently large. In such cases, the empirical marginal distributions $\hat{P}(A, B, C) $ and $\hat{P}(A, C)$ are poor estimates of the  distributions $P(A, B, C)$ and $P(A, C)$ of the model.   In our experiences, the method frequently gives negative estimates for probability values in the context of latent tree models.

In this paper, we do not estimate parameters by solving the equation in Theorem~\ref{theorem.1}. However, we adopt the progressive estimation scheme and combine it with EM. This gives rise to {\em progress EM (PEM)}. To estimate the parameters of $m_1$, PEM first estimates $P(Y)$, $P(A|Y)$, $P(B|Y)$, and $P(C|Y)$ by running EM on the sub-model $Y$-$\{A, B, C\}$; then it estimates $P(D|Y)$ by running EM on the sub-model $Y$-$\{B, C, D\}$ with $P(B|Y)$, $P(C|Y)$ and $P(Y)$ fixed; and finally it estimates $P(E|Y)$ on sub-model $Y$-$\{C, D, E\}$ similarly. All the sub-models involve 3 observed variables. 

For $m_2$, PEM first estimates $P(Y)$, $P(A|Y)$, $P(B|Y)$ and $P(D|Y)$ by running EM on  sub-model $Y$-$\{A, B, D\}$; then it estimates $P(C|Z)$, $P(E|Z)$ and $P(Z|Y)$ by running EM on the two latent variable sub-model $\{B, D\}$-$Y$-$Z$-$\{C, E\}$, with $P(B|Y)$, $P(D|Y)$ and $P(Y)$ fixed. Note that only two of the children of $Y$ are used here, and the model involves only 4 observed variables.

 Intuitively, the moments method tries to fit data in a rigid way, while PEM tries to fit data in an elastic  manner.  It never gives negative probability values. Moreover, it is still efficient because EM is run only on sub-models with three or four observed binary variables, and local maxima is seldom an issue using multiple starting points.

\subsection{\uc{PEM for Island Building}}
\vspace{-2mm}
PEM can be aligned with the subroutine \sysname{OneIsland} nicely because the subroutine adds 
variables to the working set $\mathcal{S}$ one at a time. Consider a pass through the loop. 
At the beginning, we have an LCM $m$ for the variables in $\mathcal{S}$, whose parameters 
have been estimated earlier. Then \sysname{OneIsland} finds the variable $X$  outside
 $\mathcal{S}$  that has the maximum MI with  $\mathcal{S}$, and the variable $W$ inside  $\mathcal{S}$ that has the maximum MI with $X$ (line~\ref{alg3.pickout},~\ref{alg3.pick-x}).

At line~\ref{alg3.return-lcm}, \sysname{OneIsland} adds $X$ to the $m$  to create a new LCM $m_1$, and estimates the parameters for the new variable
using the subroutine  \sysname{PEM-LCM}.  We illustrate how this is done using Fig \ref{fig.ud-test}. Suppose the LCM $m$ is the model $Y$-$\{A, B, C, D\}$ and the variable $X$ is $E$.  
\sysname{PEM-LCM} adds the variable $E$ to $m$ and thereby creates a 
new LCM $m_1$, which is $Y$-$\{A, B, C, D, E\}$ (Fig \ref{fig.ud-test} left).
 To estimate the distribution $P(E|Y)$,  \sysname{PEM-LCM} creates a temporary model  $m'$ from $m_1$ by only keeping three observed variables: $E$ and two other 
 variables with maximum MI with $E$.  Suppose $m'$ is $Y$-$\{C, D, E\}$. \sysname{PEM-LCM} estimates the distribution $P(E|Y)$ by running EM on $m'$ with all other parameters fixed. Finally, it copies  $P(E|Y)$ from $m'$ to $m_1$, and returns $m_1$.

At line~\ref{alg3.ltm},  \sysname{OneIsland} 
adds $X$ to $m$ and learns a two-latent variable model $m_2$ using the subroutine
 \sysname{PEM-LTM-2L}. We illustrate  \sysname{PEM-LTM-2L} using the foregoing example.
 Let $X$ be $E$ and $W$ be $C$. \sysname{PEM-LTM-2L}
 creates the new model $m_2$, which is $\{A, B, D\}$-$Y$-$Z$-$\{C, E\}$
 (Fig \ref{fig.ud-test} right). To estimate the parameters
  $P(C|Z)$, $P(E|Z)$ and $P(Z|Y)$, 
  \sysname{PEM-LTM-2L} creates a temporary model $m'$ which is $\{A, D\}$-$Y$-$Z$-$\{C,  E\}$.
  Only the two of the children of $Y$ that have maximum MI with $E$ remain($A$ and $D$ in this example).
  \sysname{PEM-LTM-2L} estimates 
  the three distributions  by running EM on $m'$ with all other parameters fixed.
  Finally, it copies the distributions
 from $m'$ to $m_2$ and  returns $m_2$.  \footnote[1]{Details of \sysname{PEM-LCM} and \sysname{PEM-LTM-2L} can be found in the Appendix submitted as a supplement.} Similarly in the subroutine \sysname{BridgedIslands} we use this method to estimate  parameters for edges between latent variables, but only estimating $P(Z|Y)$ and keeping all other parameters fixed.

\section{\uc{Empirical results}} \label{sec.results}
\vspace{-3mm}
We aim at scaling up HLTA, hence we need to empirically determine how efficient PEM-HLTA is compared with HLTA. We also compare PEM-HLTA with nHDP, the state-of-the-art LDA-based method for hierarchical topic detection, in terms of computational efficiency and quality of results. Also included in the comparisons are hLDA and a method named CorEx~\citep{versteeg14discovering} that builds hierarchical latent trees by optimizing an information-theoretic objective function.

Two  of the datasets used  are NIPS data\footnote[2]{http://www.cs.nyu.edu/~roweis/data.html} and Newsgroup\footnote[3]{http://qwone.com/˜jason/20Newsgroups/}. Three versions of the NIPS data with vocabulary sizes 1,000, 5,000 and 10,000 were created by choosing words with highest average TF-IDF values, referred to as Nips-1k, Nips-5k and Nips-10k. Similarly, two versions (News-1k and News-5k) of the Newsgroup data were created. Note that News-10k is not included because it is beyond the capabilities of three of the methods. Comparisons of PEM-HLTA and nHDP on large-scale data will be given separately in Section ~\ref{sec.stochasEM}. After preprocessing, NIPS and Newsgroup consist of 1,955 and 19,940 documents respectively.  For PEM-HLTA, HLTA and CorEx, the data are represented as binary vectors, whereas for nHDP and hLDA, they are represented as bags-of-words.

PEM-HLTA determines the height of hierarchy and the number of nodes at each level automatically. On the NIPS and Newsgroup datasets, it produced hierarchies with between 4 to 6 levels. For nHDP and hLDA, the height of hierarchy needs to be manually set and is usually set at 3.  We set the number of nodes at each level in such way that nHDP and hLDA would yield roughly the same total number of topics as PEM-HLTA. CorEx were configured similarly. PEM-HLTA is implemented in Java. The parameter settings are described in Section~\ref{sec.algorithm}. Implementations of other algorithms were provided by their authors and ran at their default parameter settings. All experiments are conducted on the same desktop computer.

\subsection{\uc{Topic Hierarchies for Nips-10k}}\label{sec.TopicExtract}
Table \ref{tbl:topics} shows parts of the topic hierarchies obtained by nHDP and PEM-HLTA. The left half displays 3 top-level topics by nHDP and their children. Each nHDP topic is represented using the top 5 words occurring with highest probabilities in the topic. The right half show 3 top-level topics yielded by PEM-HLTA and their children.  The topics are extracted from the model learned by PEM-HLTA as follows:
For a latent binary variable $Z$ in the model, we enumerate
the word variables in the sub-tree rooted at Z in descending order of their MI values with Z. The
leading words are those whose probabilities differ the most between the two states of $Z$ and are hence
used to characterize the states. The state of  $Z$ under which the words occur less often overall is
regarded as the {\em background topic} and is not reported, while the other state is reported as a genuine  topic. Values in [] show the percentage of the documents belonging to the genuine topic.

Let us examine some of the topics. We refer to topics on the left using the letter ‘L’ followed by topic numbers and
those on the right using ‘R’. For PEM-HLTA, R1 consists of probability terms: R1.1 is about EM algorithm; R1.2 about Gaussian mixtures and R1.3 about generative distributions.  R1.4 is a combination of variance and noise, which are separated at the next lower level. For nHDP, the topic L1 and its children L1.1, L1.2 and L1.5 are also about probability. However, L1.3 and L1.4 do not fit in the group well.
The topic R2 is about image analysis, while its first four subtopics are about different aspects of image analysis: \emph{sources of images, pixels, objects}. R2.5 and R2.6 are also meaningful and related, but do not fit in well. They are placed in another subgroup by PEM-HLTA. In nHDP, the subtopics of L2 do not give a clear spectrum of aspects of image analysis. The topic R3 is about speech recognition. Its subtopics are about different aspects of speech recognition. Only R3.4 does not fit in the group well.  In contrast, L3 and its subtopics do not present a clear semantic hierarchy.  Some of them are not meaningful. Another topic related to speech recognition L1.5 is placed elsewhere. Overall, the topics and topic hierarchy obtained by PEM-HLTA are more meaningful than those by nHDP.

\begin{table*}[t]
\vspace{-3mm}
\centering
\caption{Parts of the topic hierarchies obtained by nHDP (left) and PEM-HLTA (right) on Nips-10k.}
\vspace{-1mm}
{\tt \begin{scriptsize}

\begin{tabular}{cc}

{	
\hspace{-1cm}	\begin{tabular}{p{6.5cm}}

{\bf 1.	gaussian likelihood mixture density Bayesian }\\
  \quad 1.1. gaussian density likelihood	Bayesian \\ 
  \quad 1.2. frey hidden	posterior chaining log \\
  \quad 1.3. classifier classifiers confidence \\ 
  \quad 1.4. smola adaboost onoda mika svms \\
  \quad 1.5. speech context hme hmm	experts\\  \\

{\bf 2.	image recognition images feature features }\\
  \quad 2.1. image	recognition feature images object \\
  \quad 2.2. smola adaboost	onoda utterance \\ 
  \quad 2.3. object matching	shape image features	\\
  \quad 2.4. nearest basis examples rbf	classifier	\\
  \quad 2.5. tangent distance	simard distances \\
	\\
{\bf 3.	rules language rule sequence context}  \\
  \quad 3.1. recognition speech mlp word trained		\\
  \quad 3.2. rules rule stack machine examples	 \\
  \quad 3.3. voicing syllable fault faults	units \\
  \quad 3.4. rules hint table	hidden structure	\\
  \quad 3.5. syllable stress nucleus	heavy bit  \\
\\

    \end{tabular}

 }
 &\hspace{-1cm}
 {

    \begin{tabular}{p{8cm}}
 {\bf    1.	[0.22] mixture gaussian mixtures em covariance} \\
  \quad 1.1.  [0.23]  em maximization ghahramani expectation \\ 
  \quad \underline{1.2.  [0.23]  mixture} gaussian mixtures covariance \\ 
  \quad 1.3.  [0.23]  generative dis generafive generarive \\ 
  \quad 1.4.  [0.27]  variance noise exp variances deviation \\
     \quad   \quad   1.4.1. [0.28] variance exp variances deviation cr \\
  \quad    \quad     1.4.2. [0.44] noise noisy robust robustness mea \\
\\

{\bf 2.	[0.26] images image pixel pixels object} \\
\quad 2.1.  [0.25] images image features detection face \\
\quad 2.2.  [0.24] camera video imaging false tracked \\
\quad 2.3.  [0.24] pixel pixels intensity intensities \\ 
\quad \underline{2.4.   [0.17] object} objects shape views plane  \\
\quad 2.5.  [0.20] rotation invariant translation \\
\quad 2.6.  [0.26] nearest neighbor kohonen neighbors \\ 
\\

{\bf 3.	[0.15] speech word speaker language phoneme }\\
\quad  3.1.  [0.16] word language vocabulary words sequence \\
\quad  3.2.  [0.11] spoken acoustics utterances speakers \\
\quad  3.3.  [0.10] string strings grammar symbol symbols \\
\quad  \underline{3.4.  [0.06] retrie}val search semantic searching  \\ 
\quad  3.5.  [0.14] phoneme phonetic phonemes waibel lang \\
\quad  3.6.  [0.15] speech speaker acoustic hmm hmms \\ \\

    \end{tabular}

}

  \end{tabular}
\end{scriptsize}
}
\label{tbl:topics}
\vspace{-1cm}
\end{table*}
\vspace{-3mm}
\subsection{\uc{Topic Coherence and Model Quality}}
\vspace{-2mm}
To quantitatively measure the quality of the topics, we use the 
 \emph{topic coherence score} proposed by  ~\cite{mimno2011optimizing}. The metric depends on the number $M$ of words used to characterize a topic. We set $M=4$.  In addition, we use held-out likelihood to assess the quality of the models produced by the five algorithms. Each dataset was randomly partitioned into a training set with  80\% of the data, and a test set with  20\% of the data.  

Table~\ref{tbl:TopicCoherence} shows the average topic coherence scores of the topics produced by the five algorithms.  The sign ``-'' indicates running time exceeded 72 hours. The quality of topics produced by PEM-HLTA is similar to those by HLTA on Nips-1k and News-1k, and better on Nips-5k. In all cases, PEM-HLTA produced significantly better topics than nHDP and the other two algorithms. 
The held-out per-document loglikelihood statistics are shown in Table~\ref{tbl:perll}. The likelihood values of PEM-HLTA are similar to those  of HLTA, showing that the use of PEM to replace EM does not influence model quality much. They are significantly higher than those of CorEx.  Note that the likelihood values in Table~\ref{tbl:perll} for the LDA-based methods are calculated from bag-of-words data. They are still lower than the other methods even calculated from the same binary data as for the other three methods. 

It should be noted that, in general,  better model fit does not necessarily imply better topic quality~\citep{chang2009reading}. In context of hierarchical topic detection, however, PEM-HLTA not only leads to better model fit, but also gives better topics and better topic hierarchies.
\vspace{-4mm}
\begin{minipage}{\textwidth}
\begin{minipage}{.5\textwidth}
{\rowcolors{2}{white}{gray!20}
\begin{table}[H]
\tabcolsep=0.05cm
\caption{Average topic coherence scores.}
\vspace{-4mm}
\begin{small}
	\begin{center}
	\begin{tabular}{l|ccccc}
        &	{\scriptsize{Nips-1k}}&	{\scriptsize{Nips-5k}}&	{\scriptsize{Nips-10k}}&	 {\scriptsize{News-1k}}&	{\scriptsize{News-5k}} \\ \hline
	{\scriptsize{PEM-HLTA}}	& -6.25&{\bf -8.04}&{\bf -8.87}&-12.30&{\bf -13.07} \\ 
    HLTA   &{\bf -6.23} & -9.23&--- & {\bf-12.08}&---\\ 
    hLDA   & -6.99 &-8.94&--- & ---&--- \\ 
    nHDP  & -8.08& -9.55 &-9.86 & -14.26& -14.51\\ 
    CorEx  & -7.23& -9.85& -10.64&-13.47 & -14.51 \\ 
    \end{tabular}
	\end{center}
\end{small}
\label{tbl:TopicCoherence}
\end{table}
}
\end{minipage}
\begin{minipage}{.5\textwidth}
{\rowcolors{2}{white}{gray!20}
\begin{table}[H]
\tabcolsep=0.08cm
\caption{Per-document loglikelihood}
\vspace{-4mm}
\begin{small}
	\begin{center}
	\begin{tabular}{l|rrrrr}
        &	{\scriptsize{Nips-1k}}&	{\scriptsize{Nips-5k}}&	{\scriptsize{Nips-10k}}&	 {\scriptsize{News-1k}}&	{\scriptsize{News-5k}} \\ \hline
	{\scriptsize{PEM-HLTA}}	& {\bf -390} &{\bf -1,117}& {\bf -1,424}& {\bf -116}&{\bf -262}\\ 
    HLTA   &-391 &-1,161&--- & -120&---\\ 
    hLDA   &-1,520 &-2,854 & --- &---&--- \\ 
  	nHDP & -3,196&-6,993 &-8,262 & -265 & -599\\ 
    CorEx  &-442 &-1,226 & -1,549&-140 & -322\\ 
    \end{tabular}
	\end{center}
\end{small}
\label{tbl:perll}
\end{table}
}
\end{minipage}
\end{minipage}

\vspace{1mm}
\subsection{\uc{Running times}}
\vspace{-1mm}

 Table~\ref{tbl:Time} shows the running time statistics.  PEM-HLTA drastically outperforms HLTA, and the difference increases with vocabulary size. On Nips-10k and News-5k, HLTA did not terminate in 3 days, while PEM-HLTA finished the computation in  about 6 hours. PEM-HLTA is also faster than nHDP, although the difference decreases with vocabulary size as nHDP works in a stochastic way \citep{paisley2012nested}. Moreover,  PEM-HLTA is  more efficient than hLDA and CorEx.

\begin{minipage}{\textwidth}
\vspace{-3mm}
\begin{small}
\begin{minipage}{.5\textwidth}
{\rowcolors{2}{white}{gray!20}
 \begin{table}[H]
 \tabcolsep=0.05cm
	\caption{Running times.}
	\vspace{-3mm}
	\begin{tabular}{l|rrrrr}
    Time(min)     &	{\scriptsize{Nips-1k}}&	{\scriptsize{Nips-5k}}&	 {\scriptsize{Nips-10k}}&	 {\scriptsize{News-1k}}&	{\scriptsize{News-5k}} \\ \hline
	{\scriptsize{PEM-HLTA}}	& {\bf 4} &{\bf 140} &{\bf 340}& {\bf 47 }& {\bf 365}\\ 
    HLTA   &42 &  2,020 &---& 279&---\\ 
    hLDA   & 2,454&4,039 &--- &--- &--- \\ 
    nHDP  &359 & 382&435 & 403&477 \\ 
    CorEx  & 43&366 &704 &722 &4,025 \\ 
    \end{tabular}
    \label{tbl:Time}
 \end{table}
}
\end{minipage}
\begin{minipage}{.5\textwidth}
{\rowcolors{2}{white}{gray!20}
 \begin{table}[H]
 \tabcolsep=0.03cm
\caption{Performances on the New York Times data.}
\vspace{6mm}
    	\label{tbl:nyt}
    	\begin{tabular}{l|r|r}
                   & Time (min) & Average topic coherence \\ \hline
         PEM-HLTA   &  670 &-12.86 \\
         hHDP        &637 &-13.35 \\
        \end{tabular}  
\end{table}
}

\end{minipage}
\end{small}
\end{minipage}

\subsection{\uc{Stochastic EM}}\label{sec.stochasEM}
\vspace{-2mm}
Conceptually, PEM-HLTA has two phases: hierarchical model construction and parameter estimation. In the second phase, EM is run a predefined number of steps from the initial parameter values from the first phase. It is time-consuming if the sample size is large.~\cite{paisley2012nested} faced a similar problem with nHDP.  They solve the problem using stochastic inference. The idea is to divide the data set into subsets and process the subsets one by one.  Model parameters are updated after processing each data subset and overall one goes through the entire data set only once.We adopt the same idea for the second phase of PEM-HLTA and call it {\em stochastic EM}. We tested the idea on the New York Times dataset\footnote[1]{http://archive.ics.uci.edu/ml/datasets/Bag+of+Words}, which consists of 300,000 articles. To analyze the data, we picked 10,000 words using TF-IDF and then randomly divided the dataset into 50 equal-sized subsets. We used only the fist subset for the first phase of PEM-HLTA. For the second phase, we ran EM on current model once using each subset in turn until all the subsets are utilized . 

On New York Times data, we only compare PEM-HLTA with nHDP since other methods are not amenable to processing large datasets as we can observe from Table~\ref{tbl:Time}. We still trained nHDP model using documents in bag-of-words form and PEM-HLTA using documents as binary vectors of words. Table~\ref{tbl:nyt} reports the running times and topic coherence.  PEM-HLTA  took around 11 hours which is a little bit slower than nHDP (10.5 hours). However,  PEM-HLTA produced more coherent topics, which is not only testified by the coherence score, but also the resulting topic hierarchies. The reader could get a clear picture of the superiority of PEM-HLTA over nHDP by taking a quick look at the model structure and topic hierarchies submitted as supplements.
\vspace{-1mm}

\section{\uc{Conclusions}} \label{sec.conclusions}
\vspace{-4mm}
We have proposed and investigated a method to scale up HLTA --- a newly emerged method for hierarchical topic detection. The key idea is to replace EM using progressive EM.  The resulting algorithm PEM-HLTA  reduces the computation time of HLTA drastically and can handle much larger datasets. More importantly, it outperforms nHDP, the state-of-the-art LDA-based method for hierarchical topic detection, in terms of both  quality of topics and topic hierarchy, with comparable speed on large-scale data. Although we only show how PEM works in HLTA, PEM can possibly be used in other more general models. PEM-HLTA can also be further scaled up through parallelization and used for text classification. We plan to investigate these directions in the future.
\newpage
\begin{small}
\bibliographystyle{plainnat}
\bibliography{us,others}

\begin{thebibliography}{24}
\providecommand{\natexlab}[1]{#1}
\providecommand{\url}[1]{\texttt{#1}}
\expandafter\ifx\csname urlstyle\endcsname\relax
  \providecommand{\doi}[1]{doi: #1}\else
  \providecommand{\doi}{doi: \begingroup \urlstyle{rm}\Url}\fi

\bibitem[Anandkumar et~al.(2012)Anandkumar, Chaudhuri, Hsu, Kakade, Song, and
  Zhang]{anandkumar12spectral}
Animashree Anandkumar, Kamalika Chaudhuri, Daniel Hsu, Sham~M. Kakade, Le~Song,
  and Tong Zhang.
\newblock Spectral methods for learning multivariate latent tree structure.
\newblock In \emph{Advances in Neural Information Processing Systems}, pages
  2025--2033, 2012.

\bibitem[Bartholomew and Knott(1999)]{bartholomew99latent}
David~J. Bartholomew and Martin Knott.
\newblock \emph{Latent Variable Models and Factor Analysis}.
\newblock Arnold, 2nd edition, 1999.

\bibitem[Blei and Lafferty(2006)]{blei2006dynamic}
David~M Blei and John~D Lafferty.
\newblock Dynamic topic models.
\newblock In \emph{Proceedings of the 23rd international conference on Machine
  learning}, pages 113--120. ACM, 2006.

\bibitem[Blei and Lafferty(2007)]{blei2007correlated}
David~M Blei and John~D Lafferty.
\newblock A correlated topic model of science.
\newblock \emph{The Annals of Applied Statistics}, pages 17--35, 2007.

\bibitem[Blei et~al.(2003)Blei, Ng, and Jordan]{blei03latent}
David~M. Blei, Andrew~Y. Ng, and Michael~I. Jordan.
\newblock Latent {D}irichlet allocation.
\newblock \emph{Journal of Machine Learning Research}, 3:\penalty0 993--1022,
  2003.

\bibitem[Blei et~al.(2010)Blei, Griffiths, and Jordan]{blei10nested}
David~M. Blei, Thomas~L. Griffiths, and Michael~I. Jordan.
\newblock The nested {C}hinese restaurant process and {B}ayesian nonparametric
  inference of topic hierarchies.
\newblock \emph{Journal of the ACM}, 57\penalty0 (2):\penalty0 7:1--7:30, 2010.

\bibitem[Chang et~al.(2009)Chang, Boyd-Graber, Gerrish, Wang, and
  Blei]{chang2009reading}
Jonathan Chang, Jordan~L Boyd-Graber, Sean Gerrish, Chong Wang, and David~M
  Blei.
\newblock Reading tea leaves: How humans interpret topic models.
\newblock In \emph{Advances in Neural Information Processing Systems},
  volume~22, pages 288--296, 2009.

\bibitem[Chang(1996)]{chang96full}
Joseph~T. Chang.
\newblock Full reconstruction of {M}arkov models on evolutionary trees:
  Identifiability and consistency.
\newblock \emph{Mathematical Biosciences}, 137\penalty0 (1):\penalty0 51--73,
  1996.

\bibitem[Chen et~al.(2012)Chen, Zhang, Liu, Poon, and Wang]{chen12model}
Tao Chen, Nevin~L. Zhang, Tengfei Liu, Kin~Man Poon, and Yi~Wang.
\newblock Model-based multidimensional clustering of categorical data.
\newblock \emph{Artificial Intelligence}, 176:\penalty0 2246--2269, 2012.

\bibitem[Chow and Liu(1968)]{chow68approximating}
C.~K. Chow and C.~N. Liu.
\newblock Approximating discrete probability distributions with dependence
  trees.
\newblock \emph{{IEEE} Transactions on Information Theory}, 14\penalty0
  (3):\penalty0 462--467, 1968.

\bibitem[Cover and Thomas(2012)]{cover2012elements}
Thomas~M Cover and Joy~A Thomas.
\newblock \emph{Elements of information theory}.
\newblock John Wiley \& Sons, 2012.

\bibitem[Dempster et~al.(1977)Dempster, Laird, and Rubin]{dempster77maximum}
Arthur~P. Dempster, Nan~M. Laird, and Donald~B. Rubin.
\newblock Maximum likelihood from incomplete data via the {EM} algorithm.
\newblock \emph{Journal of the Royal Statistical Society. Series B
  (Methodological)}, 39\penalty0 (1):\penalty0 1--38, 1977.

\bibitem[Liu et~al.(2013)Liu, Zhang, Chen, Liu, Poon, and Wang]{liu13greedy}
Teng-Fei Liu, Nevin~L. Zhang, Peixian Chen, April~Hua Liu, Leonard~K.M. Poon,
  and Yi~Wang.
\newblock Greedy learning of latent tree models for multidimensional
  clustering.
\newblock \emph{Machine Learning}, 98\penalty0 (1--2):\penalty0 301--330, 2013.

\bibitem[Liu et~al.(2014)Liu, Zhang, and Chen]{liu14hierarchical}
Tengfei Liu, Nevin~L. Zhang, and Peixian Chen.
\newblock Hierarchical latent tree analysis for topic detection.
\newblock In \emph{Machine Learning and Knowledge Discovery in Databases},
  pages 256--272, 2014.

\bibitem[Mimno et~al.(2011)Mimno, Wallach, Talley, Leenders, and
  McCallum]{mimno2011optimizing}
David Mimno, Hanna~M Wallach, Edmund Talley, Miriam Leenders, and Andrew
  McCallum.
\newblock Optimizing semantic coherence in topic models.
\newblock In \emph{Proceedings of the Conference on Empirical Methods in
  Natural Language Processing}, pages 262--272. Association for Computational
  Linguistics, 2011.

\bibitem[Paisley et~al.(2012)Paisley, Wang, Blei, and
  Jordan]{paisley2012nested}
John Paisley, Chong Wang, David~M Blei, and Michael~I Jordan.
\newblock Nested hierarchical dirichlet processes.
\newblock \emph{IEEE Transactions on Pattern Analysis and Machine
  Intelligence}, 37, 2012.

\bibitem[Pearl(1988)]{pearl88probabilistic}
Judea Pearl.
\newblock \emph{Probabilistic Reasoning in Intelligent Systems: Networks of
  Plausible Inference}.
\newblock Morgan Kaufmann Publishers, San Mateo, California, 1988.

\bibitem[Raftery(1995)]{raftery95bayesian}
Adrian~E. Raftery.
\newblock Bayesian model selection in social research.
\newblock \emph{Sociological Methodology}, 25:\penalty0 111--163, 1995.

\bibitem[Schwarz(1978)]{schwarz78estimating}
Gideon Schwarz.
\newblock Estimating the dimension of a model.
\newblock \emph{The Annals of Statistics}, 6\penalty0 (2):\penalty0 461--464,
  1978.

\bibitem[Ver~Steeg and Galstyan(2014)]{versteeg14discovering}
Greg Ver~Steeg and Aram Galstyan.
\newblock Discovering structure in high-dimensional data through correlation
  explanation.
\newblock In \emph{Advances in Neural Information Processing Systems 27}, pages
  577--585, 2014.

\bibitem[Zhang(2004)]{zhang04hierarchical}
Nevin~L. Zhang.
\newblock Hierarchical latent class models for cluster analysis.
\newblock \emph{Journal of Machine Learning Research}, 5:\penalty0 697--723,
  2004.

\bibitem[Zhang et~al.(2008{\natexlab{a}})Zhang, Wang, and Chen]{zhang08blatent}
Nevin~L. Zhang, Yi~Wang, and Tao Chen.
\newblock Latent tree models and multidimensional clustering of categorical
  data.
\newblock Technical Report HKUST-CS08-02, The Hong Kong Univeristy of Science
  and Technology, 2008{\natexlab{a}}.

\bibitem[Zhang et~al.(2008{\natexlab{b}})Zhang, Wang, and
  Chen]{zhang08discovery}
Nevin~L. Zhang, Yi~Wang, and Tao Chen.
\newblock Discovery of latent structures: Experience with the {C}o{IL}
  challenge 2000 data set.
\newblock \emph{Journal of Systems Science and Complexity}, 21:\penalty0
  172--183, 2008{\natexlab{b}}.

\bibitem[Zhang et~al.(2014)Zhang, Wang, and Chen]{zhang14study}
Nevin~L. Zhang, Xiaofei Wang, and Peixian Chen.
\newblock A study of recently discovered equalities about latent tree models
  using inverse edges.
\newblock In \emph{Probabilistic Graphical Models}, pages 567--580, 2014.

\end{thebibliography}
\end{small}

\end{document}